\documentclass[conference]{IEEEtran}
\IEEEoverridecommandlockouts

\usepackage[pdftex]{graphicx}
\usepackage{amssymb}
\usepackage{amsmath}
\usepackage{hyperref}
\usepackage{bm}
\usepackage{cite}
\usepackage{booktabs}
\usepackage{color}
\usepackage{url}
\usepackage{longtable}
\usepackage{multirow}
\usepackage{array}
\usepackage[export]{adjustbox}

\newcolumntype{x}[1]{>{\centering\arraybackslash}p{#1}}

\def\BibTeX{{\rm B\kern-.05em{\sc i\kern-.025em b}\kern-.08em
    T\kern-.1667em\lower.7ex\hbox{E}\kern-.125emX}}
\begin{document}
\title{Target-Oriented Deformation of Visual-Semantic Embedding Space\\
}

\author{\IEEEauthorblockN{Takashi Matsubara}
\IEEEauthorblockA{\textit{Graduate School of System Informatics, Kobe University} \\
1-1 Rokkodai, Nada, Kobe, Hyogo, 657-8501 Japan. \\
Email: matsubara@phoenix.kobe-u.ac.jp}
}
\maketitle

\begin{abstract}
Multimodal embedding is a crucial research topic for cross-modal understanding, data mining, and translation.
Many studies have attempted to extract representations from given entities and align them in a shared embedding space.
However, because entities in different modalities exhibit different abstraction levels and modality-specific information, it is insufficient to embed related entities close to each other.
In this study, we propose the Target-Oriented Deformation Network (TOD-Net), a novel module that continuously deforms the embedding space into a new space under a given condition, thereby adjusting similarities between entities.
Unlike methods based on cross-modal attention, TOD-Net is a post-process applied to the embedding space learned by existing embedding systems and improves their performances of retrieval.
In particular, when combined with cutting-edge models, TOD-Net gains the state-of-the-art cross-modal retrieval model associated with the MSCOCO dataset.
Qualitative analysis reveals that TOD-Net successfully emphasizes entity-specific concepts and retrieves diverse targets via handling higher levels of diversity than existing models.
\end{abstract}

\begin{IEEEkeywords}
  visual-semantic embedding, image-caption retrieval, flow-based model
\end{IEEEkeywords}

\section{Introduction}
Our society is immersed in the age of big data, which includes diverse  modalities, such as text, images, audio, and video.
Entities in different modalities exhibit different abstraction levels and modality-specific information.
This property increases the difficulty of cross-modal understanding and data mining.
Existing works have mainly employed multimodal embedding, which maps entities in different modalities to vectors in a common space~\cite{Frome2013,Kiros2014,Zheng2017,Peng2017,Gu2017,Faghri2018,Engilberge2018}.
Euclidean measures then evaluate the similarity between entities in order to facilitate tasks such as retrieval and translation.

\begin{figure}[t]\centering
  \includegraphics[page=1,scale=0.5,clip]{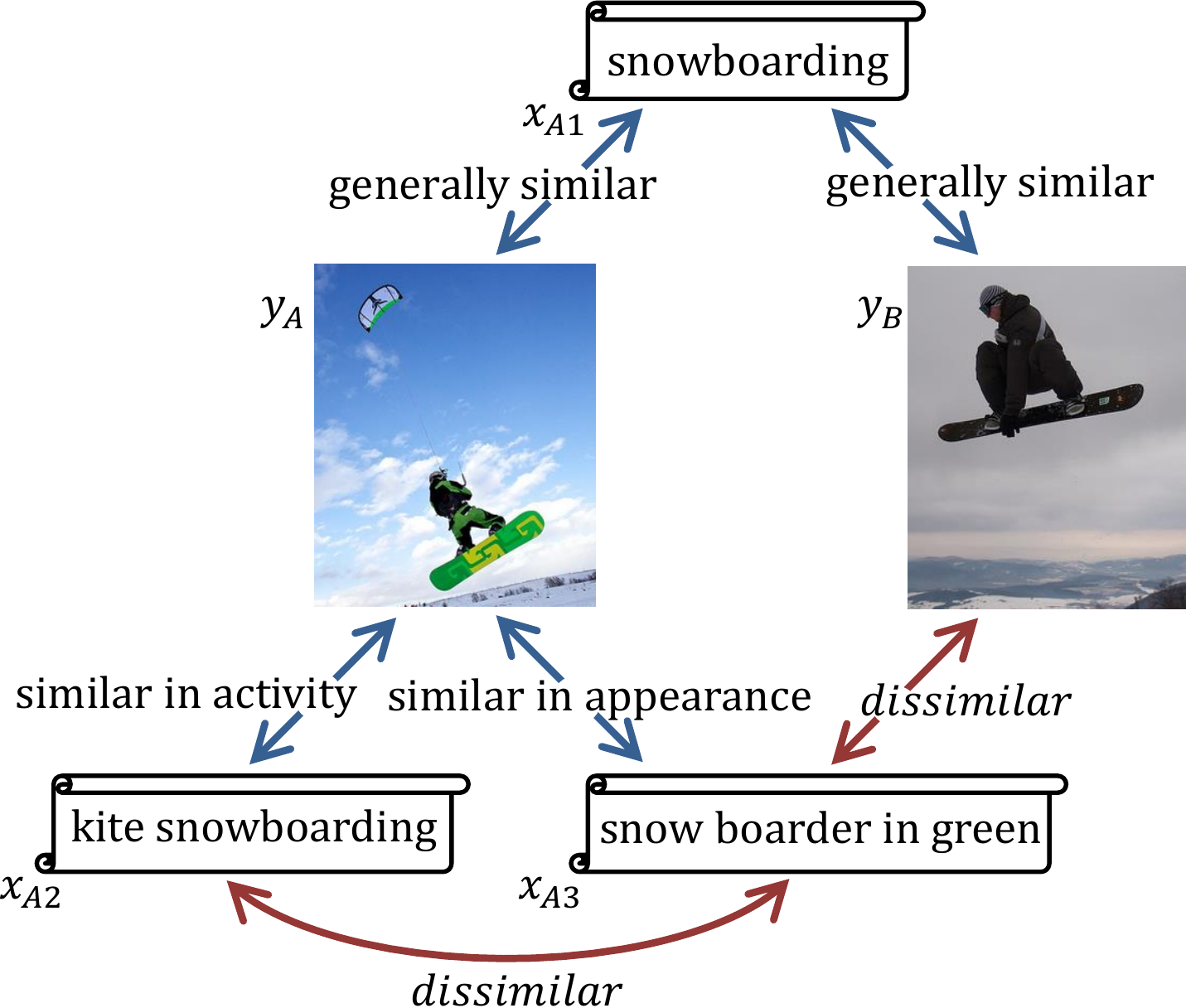}
  \caption{Conceptual diagram of the problem of interest in this study.
  The captions $x_{A1}$, $x_{A2}$, and $x_{A3}$ reference specific aspects of image $y_A$, and hence, they are dissimilar.
  The caption $x_{A1}$ can match the other image $y_B$, but the other captions should not.
  Therefore, it is insufficient to embed them close to each other in the Euclidean space.
  }
  \label{fig:concept}
\end{figure}

However, as shown in Fig.~\ref{fig:concept}, Euclidean embedding has a limitation.
Let us focus on visual-semantic embedding, and imagine an image $y_A$ that depicts a person dressed in green and engaging in kite snowboarding.
A caption $x_{A1}$ simply describes the activity as snowboarding.
Meanwhile, other images could also be retrieved, such as an image $y_B$ depicting a snowboarder dressed in black.
Another caption $x_{A2}$ specifically describes the activity as kite snowboarding, and can retrieve image $y_A$ selectively.
Yet another caption $x_{A3}$ focuses on the appearance (green clothing) and retrieves image $y_A$.
The latter two captions $x_{A2}$ and $x_{A3}$ are dissimilar and should not retrieve image $y_B$.
Captions often reference a specific aspect of an image; hence, it is insufficient to embed these entities close to each other.

Several studies have employed an ordered vector space or hyperbolic space to capture the hierarchical relationship between words (hypernym and hyponym) or tree nodes~\cite{Vendrov2015,Nickel2017,Athiwaratkun2018}.
Conversely, visual-semantic embedding has only two hierarchies (caption and image) and cannot benefit from the constraints of hierarchial relationships.
In recent studies that focused specifically on image-caption retrieval, embedding was replaced with a cross-modal attention to directly obtain similarity scores~\cite{Lee2018,Wang2019b}.
Despite the improved performance, their generality to other tasks and modalities is limited.
Recently, in the context of fashion recommendations, a conditional similarity network (CSN) was proposed~\cite{Veit2017,Tan2019}.
For retrieval, the CSN focuses on a target aspect (e.g., color) and ignores others (e.g., category and occasion) by disregarding dimensions of no interest.
Similarly, manipulating the Euclidean space for visual-semantic embedding potentially overcomes the gap between the modalities.

In this paper, we propose the \emph{Target-Oriented Deformation Network} (TOD-Net).
TOD-Net is constructed using a flow-based model, namely a conditional version of Real-NVP network~\cite{Dinh2016}.
In flow-based models, only continuous bijective functions are approximated theoretically.
TOD-Net is installed on top of an existing visual-semantic embedding system and continuously deforms the embedding space into a new space by virtue of the bijective property.
Through the deformation, the embedding space becomes specialized to a specific concept in the condition and adjusts for the similarity accordingly.
For example, if the condition is a caption describing the appearance, TOD-Net emphasizes the concept describing the appearance in the embedded images and avoids false positives that show other appearance aspects.

To end the Introduction, we summarize the contributions of this paper.
\begin{itemize}
 \item We TOD-Net, which is installed on top of a visual-semantic embedding system.
 TOD-Net learns a conditional bijective mapping and deforms the shared embedding space into a new space.
 By that means, TOD-Net adjusts the similarities between entities under a condition while preserving the topological relations between them.
 \item Unlike existing methods based on an object detector and a cross-modal attention~\cite{Huang2018f,Lee2018,Wang2019b,Liu2019,Lee2019,Huang2019b}, TOD-Net is applied to a fixed embedding space and improves the retrieval performance even when using a single-image encoder.
 This fact indicates that a single-image encoder already extracts detailed concepts from entities but encounters a difficulty in expressing their relations in the embedding space.
 Since the object detector is not needed and TOD-Net is used at the very last phase, the computational cost is greatly suppressed.
 \item We TOD-Net with existing models and conduct extensive experiments.
 The numerical results demonstrate that TOD-Net generally improves the performance of existing models that are based on visual-semantic embedding, thus achieving a state-of-the-art model for image-caption retrieval.
 \item A qualitative analysis demonstrates that TOD-Net successfully captures entity-specific concepts, which are often suppressed by existing models because of the diversity among entities belonging to the same group.
\end{itemize}


\section{Related work}\label{sec:related}

\subsection{Conventional and hierarchical embedding}
An embedding system maps entities such as words, text, and images to vectors in an Euclidean embedding space.
The similarity between two entities is defined through negative Euclidean distance, inner product, or cosine similarity in the embedding space.
Numerous studies on  visual-semantic embedding have investigated network architectures and objective functions.
Typically, a pretrained convolutional neural network (CNN) has been employed for encoding images~\cite{Frome2013,Kiros2014,Faghri2018,Engilberge2018,Engilberge2019}.
For captions, a recurrent neural network (RNN) following a word embedding model has been a common choice~\cite{Mikolov2013a}.

Captions often focus on a specific aspect of an image while ignoring others, as depicted in Fig.~\ref{fig:concept}.
This means that a caption is an abstract entity of an image and there exists a hierarchical relationship between them.
Words and graphs also have hierarchical relationships between hypernym and hyponym or between a parent node and its children in a tree.
Hence, point embedding to the Euclidean space has a limitation.
Order-embedding and related works have tackled this difficulty.
They embedded an entity as a vector in an ordered vector space~\cite{Vendrov2015}, a vector or a cone in a hyperbolic space~\cite{Nickel2017,Ganea2018}, and a Gaussian distribution~\cite{Sun2018,Athiwaratkun2018}.
These studies have embed entities so that two hyponyms are less similar to each other than to their hypernym, and have exhibited remarkable results in the word and graph embeddings.
However, visual-semantic embedding has only two hierarchies (image and caption) and cannot benefit from the constraints of hierarchical relationships.
In the original study on order-embedding, entities were embedded in a super sphere for the visual-semantic embedding even though such embedding cannot express hierarchical relationships~\cite{Vendrov2015,Faghri2018}.

\subsection{Adaptive embedding}
Recently, in the context of fashion recommendations, a conditional similarity network (CSN) was proposed~\cite{Veit2017}.
There are many aspects of similarity to consider when retrieving a fashion item (e.g., occasion, category, and color).
For example, white sneakers are similar to jeans from the aspect of occasion, to court shoes from the aspect of category, and to a white jacket from the aspect of color.
CSN learns a template for each aspect in order to rescale the dimensions of the embedded vectors, thereby emphasizing a given aspect.
In another example, SCE-Net~\cite{Tan2019} extended CSN by inferring an appropriate aspect from a given input pair.

As in these studies, our work adopts the concept of conditional adjustment of the embedding space.

\subsection{Cross-model attention}
Cutting-edge methods for image-caption retrieval sometimes employ an object detector and a cross-modal attention~\cite{Huang2018f,Lee2018,Wang2019b,Liu2019,Lee2019,Huang2019b}.
A pretrained object detector crops multiple subregions in an image, and then a cross-modal attention is performed over the cropped regions and words in a caption.
A cross-modal attention pays attention to a subset of the cropped regions related to the words to evaluate their similarities while discarding remaining subsets.
Thereby, the cross-modal attention handles hierarchy and polysemicity.

A main drawback of these studies is computational time.
Object detectors require an image larger than that required by single-image encoders and perform an additional region-proposal step.
Cross-modal attentions are performed for every possible pairs of regions and words, and their computational cost is proportional to the numbers of entities, regions, and words.
Conversely, TOD-Net is a tiny neural network that receives only embedded vectors; its computational cost much less than that of the cross-modal attention even though it is still higher than that of cosine similarity.
Moreover, an attention module is designed specifically for image-caption retrieval and its generality to other modalities and tasks is limited, while the output of TOD-Net is still an embedded vector that is potentially applicable to other tasks.

\begin{figure*}[t]\centering
  \includegraphics[page=2,scale=0.5,clip]{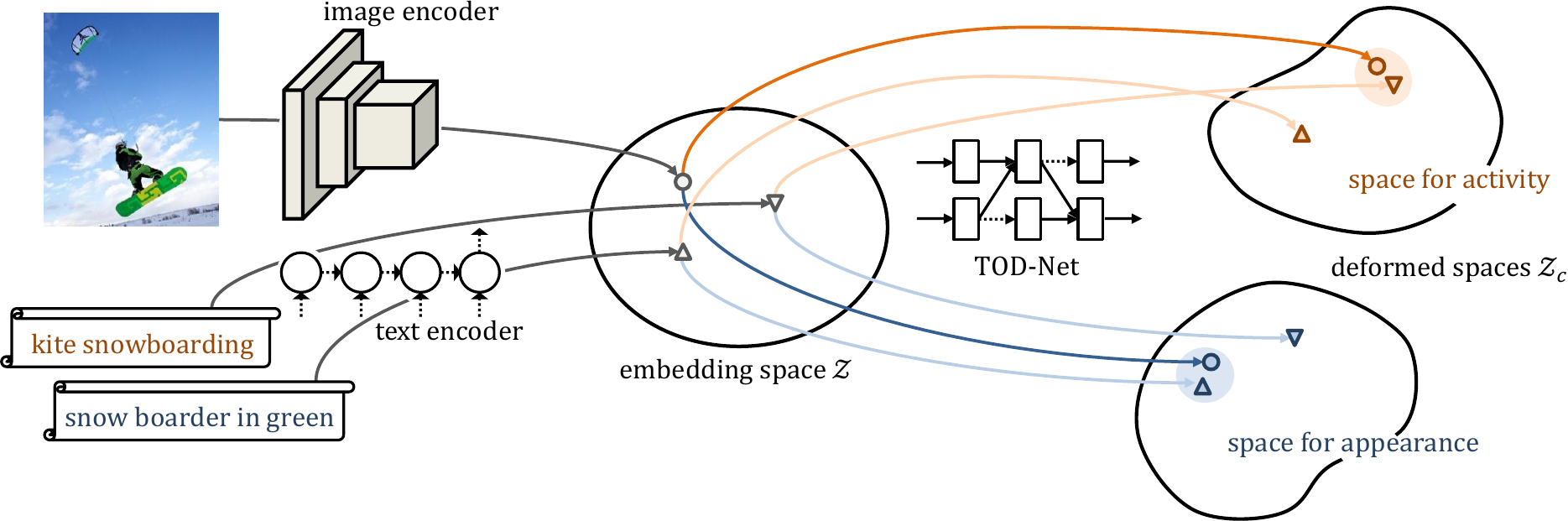}
  \caption{Conceptual diagram of the proposed \emph{Target-Oriented Deformation Network} (TOD-Net).
  A visual-semantic embedding system has an image encoder and a text decoder that map images and captions to vectors in a shared embedding space $\mathcal Z$.
  However, captions often reference a specific aspect of an image, and their hierarchical relationship is never evaluated appropriately as long as the embedding space $\mathcal Z$ is Euclidean.
  TOD-Net deforms the embedding space $\mathcal Z$ under a condition $c$ and provides new embedding spaces $\mathcal Z_c$.
  By that means, entity-specific detailed concepts such as appearance, activity, or background are emphasized, and diverse targets can be retrieved by a single query.
  }
  \label{fig:deformation}
\end{figure*}

\section{Methods}\label{sec:method}
\subsection{Preliminaries}\label{sec:prelim}
We assume that a backbone model is composed of an image encoder $E_{\mathcal X}$ and a text encoder $E_{\mathcal Y}$.
The image encoder is composed of a convolutional neural network (CNN), a pooling layer, and a fully connected layer~\cite{Faghri2018,Engilberge2018}.
When an image $x$ in an image data space $\mathcal X$ is given, the image encoder $E_{\mathcal X}$ maps the image $x$ to a vector $\tilde x$ in a $d$-dimensional embedding space $\mathcal Z$.
The text encoder $E_{\mathcal Y}$ is a recurrent neural network (RNN) or a transformer network~\cite{Devlin2018}.
It also maps a given caption $y$ in a caption data space $\mathcal Y$ to a vector $\tilde y$ in the same embedding space $\mathcal Z$.
In other words, $E_{\mathcal X}: \mathcal X\rightarrow \mathcal Z$ and $E_{\mathcal Y}: \mathcal Y\rightarrow \mathcal Z$.
Note that these maps are sometimes stochastic in the training phase due to stochastic components such as dropout and batch normalization~\cite{Ioffe2015}.
The encoders are trained under an objective function that evaluates the similarity between two entities using cosine similarity;
\begin{equation*}
    sim(x,y)
    =\frac{\langle \tilde y,\tilde x\rangle}{||\tilde y||\cdot||\tilde x||},
\end{equation*}
where $\tilde x=E_{\mathcal X}(x)$, $\tilde y=E_{\mathcal Y}(y)$.
$\langle \cdot,\cdot\rangle$ and $||\cdot||$ denote the inner product and the Euclidean norm, respectively.
Other similarities defined in a vector space are acceptable.

The quality of the embedding was evaluated using image-caption retrieval, which receives a query and then ranks targets appropriately according to the similarity~\cite{Frome2013,Kiros2014,Zheng2017,Peng2017,Gu2017,Faghri2018,Engilberge2018}.
As shown in Fig.~\ref{fig:concept}, image $y_A$ can be retrieved by two distinct captions $y_{A2}$ and $y_{A3}$.
The image $y_A$ depicts a person dressed in green and engaging in kite snowboarding.
Caption $y_{A2}$ describes the activity in detail and the other caption $y_{A3}$ focuses on the appearance; thus, each caption references a different aspect of image $y_A$.
When a model focuses on appearance, caption $y_{A3}$ comes closer to image $y_A$ in the embedding space $\mathcal Z$, while caption $y_{A2}$ becomes further away.

\subsection{Target-oriented deformation network}\label{sec:tod}
To adjust the embedded vectors, we propose a target-oriented deformation network (TOD-Net) for visual-semantic embedding.
The usage is summarized in Fig.~\ref{fig:deformation}.
TOD-Net $D_c$ is defined with a condition $c\in \mathcal Z$, which is also an embedded vector.
TOD-Net $D_c$ receives an embedded vector $v\in\mathcal Z$ and outputs a vector of the same size.
When feeding a set of embedded vectors in $\mathcal Z$, the outputs of TOD-Net $D_c$ form a new embedding space $\mathcal Z_c$ under the condition $c$, that is, $D_c:\mathcal Z\rightarrow\mathcal Z_c$.
TOD-Net $D_c$ is then trained according to the similarity in the new embedding space $\mathcal Z_c$ while the original embedding space $\mathcal Z$ is unchanged.

To obtain the similarity between a pair consisting of the query and one of the targets for image-caption retrieval, the target is fed to TOD-Net as condition $c$.
Then, the pair is mapped into the new embedding space $\mathcal Z_c$ by TOD-Net $D_c$, and its similarity is calculated as the cosine similarity in the new embedding space $\mathcal Z_c$.
For example, the similarity for a caption retrieval is
\begin{equation*}
  sim(x,y;D_c)=\frac{\langle D_{\tilde x}(\tilde y),D_{\tilde x}(\tilde x)\rangle}{||D_{\tilde x}(\tilde y)||\cdot||D_{\tilde x}(\tilde x)||},
\end{equation*}
where $x$ and $y$ denote a caption and an image, respectively, and $\tilde x$ and $\tilde y$ denote their embedded vectors.
Hence, we refer to the proposed method as target-oriented.
From a viewpoint of neural networks, TOD-Net is installed on top of an existing visual-semantic embedding system.

Through this process, TOD-Net $D_c$ is expected to deform the embedding space $\mathcal Z$ such that the concepts indicated by condition $c$ are emphasized while the others are ignored.
For example, when a caption $x$ related to appearance is fed to TOD-Net $D_c$ as a condition $c$, TOD-Net $D_c$ pays attention to the appearance in image $y$ and suppresses other concepts such as activity, weather, and background.

\begin{figure}[t]\centering
  \includegraphics[page=3,scale=0.5,clip]{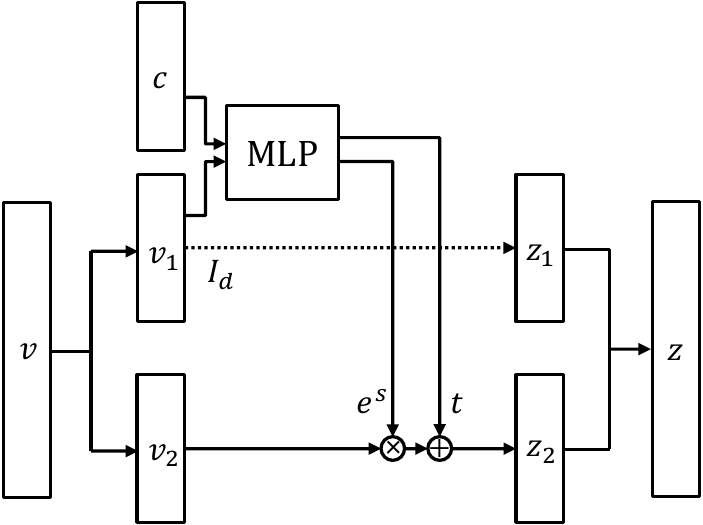}
  \caption{Diagram of the forward path of a coupling layer of the conditional Real-NVP network~\cite{Dinh2016}.
  In this paper, we do not use the backward path.
  }
  \label{fig:crealnvp}
\end{figure}

\subsection{Construction of TOD-Net}
When TOD-Net $D_c$ is a basic multilayer perceptron (MLP), it can approximate arbitrary continuous functions~\cite{Cybenko1989}.
However, such a network risks disturbing the relations between embedded vectors in the original embedding space $\mathcal Z$ that the embedding system has learned.
Instead, we employ a conditional version of the Real-NVP network to TOD-Net $D_c$~\cite{Dinh2016}, depicted in Fig.~\ref{fig:crealnvp}.
The Real-NVP network is a neural network architecture categorized into flow-based models, which only approximate continuous bijective functions theoretically.
The flow-based models have been investigated for generative models in which a sample likelihood is calculable by changing the variables.
Owing to its continuous bijective nature, TOD-Net $D_c$ continuously deforms the original embedding space $\mathcal Z$ into $\mathcal Z_c$ and is expected to adjust only similarity (distance, metric) while conserving the topological relations between embedded vectors.

The conditional Real-NVP network is composed of multiple coupling layers, each of which is an MLP in our work.
An embedded vector $v$ is divided into two vectors $v_1$ and $v_2$, which are of $\frac{d}{2}$-dimensions.
One vector $v_1$ and a condition $c$ are jointly fed to the coupling layer, which leads to the creation of two $\frac{d}{2}$-dimensional vectors $t$ and $s$.
With vectors $t$ and $s$, the remaining vector $v_2$ is linearly transformed as
\begin{equation*}
  \left\{
  \begin{split}
    z_1&=v_1\\
    z_2&=v_2 \odot \exp(s(v_1,c))+t(v_1,c),
  \end{split}
  \right.
\end{equation*}
where $\odot$ denotes the element-wise product and the input $v_1$ is kept as it is.
The pair of resultant vectors $z=\{z_1,z_2\}$ is the output of the coupling layer.
Intuitively, the vector $v_2$ is ciphered with the key vectors $v_1$ and $c$.
The coupling layer is continuous as long as the MLP is continuous.
In the following coupling layer, $z_1$ is ciphered by using the pair of $z_2$ and $c$ as a key.
Each coupling layer forms a bijective function because its inverse function can be obtained as
\begin{equation*}
  \left\{
  \begin{split}
    v_1&=z_1\\
    v_2&=(z_2-t(z_1,c)) \odot \exp (-s(z_1,c)).
  \end{split}
  \right.
\end{equation*}
The conditional Real-NVP network is a composition of coupling layers and thus it also forms a bijective function.
The hyperparameters are the number of coupling layers, the number of hidden layers in a coupling layer, and the number of units of a hidden layer.

\section{Experiments}
\subsection{Backbone models}
To evaluate our proposed TOD-Net, we employed VSE++~\cite{Faghri2018}, DSVE-loc~\cite{Engilberge2018}, and the BERT model~\cite{Devlin2018} as backbone models.
We followed the experimental settings shown in the original studies unless otherwise stated.

VSE++ is a commonly used baseline.
The image encoder is a 152-layer ResNet~\cite{He2015a,He2016a} that is pretrained using the ImageNet dataset~\cite{Deng2009}.
The final fully connected layer is replaced for embedding.
The text encoder is a single-layer GRU network~\cite{Cho2014} trained from scratch.
The dimension number of the embedding space $\mathcal Z$ is $d=1024$.
The source code can be found in the original repository\footnote{\url{https://github.com/fartashf/vsepp}.}.

DSVE-loc is a state-of-the-art model for image-caption retrieval.
The image encoder is a modified 152-layer ResNet that has a convolution layer and a special pooling layer instead of a global pooling layer~\cite{Durand2016}, and is pretrained using the ImageNet dataset~\cite{Deng2009}.
The text encoder is composed of a word embedding model pretrained by Skip-Thought~\cite{Kiros} and a four-layer SRU network~\cite{Lei2017a} trained from scratch.
The dimension number of the embedding space $\mathcal Z$ is $d=2400$.

We also report results obtained using BERT model as the text encoder.
It is based on a transformer network and is pretrained using various language processing tasks~\cite{Devlin2018}\footnote{
We used the pretrained model posted on \url{https://github.com/huggingface/pytorch-transformers}}.
After the transformer layers, the outputs are averaged over a sentence, and a fully connected layer is added for embedding.
For the image encoder, we employed a modified and pretrained ResNet~\cite{Durand2016}.
The dimension number of the embedding space $\mathcal Z$ is $d=1024$.
This model is not an existing work.

\subsection{Dataset and training procedure}
We evaluated our proposal on the MS COCO dataset~\cite{Lin2014} using the splits employed by VSE++~\cite{Faghri2018}.
We used 113,287 images for training, 5,000 images for validation, and 5,000 images for testing.
Each image has five captions as positive targets.

For image retrieval, each backbone model was trained so that the similarity between a query caption $y_q$ and a designated target image $x_p$ (called a positive target) would be larger than that to another target image $x_n$ (called a negative target).
The loss function was represented by a hinge rank loss, as follows:
\begin{equation*}
  \mathcal L(y_q,x_p,x_n)=|sim(y_q,x_n)-sim(y_q,x_p)+m|_+,
\end{equation*}
where $|\cdot|_+$ is the positive part and $m>0$ is a margin parameter.
In a mini-batch, all images except a positive target image $x_p$ are negative $\{x_n\}$.
Following VSE++~\cite{Faghri2018}, we only minimized the loss function with the hardest negative target image in a mini-batch.
Specifically,
\begin{equation}
  \mathcal L(y_q,x_p,\{x_n\})=\max_{x_n} |sim(y_q,x_n)-sim(y_q,x_p)+m|_+.
\end{equation}
The loss function for caption retrieval was defined in the same way.
The similarity function was the cosine similarity.
The batch size was 128 for VSE++ and 160 for DSVE-loc and the BERT model.

Each backbone model was trained using the Adam optimizer with the hyperparameters $\beta_1=0.9$ and $\beta_2=0.999$~\cite{Kingma2014b}.
Note that the source code of VSE++ in the original repository was updated; we trained it from scratch.
We trained its text encoder and the embedding layer of its image encoder for 30 epochs and the whole model for 15 epochs.
The learning rate was initialized to $\alpha=2\times 10^{-4}$ and then multiplied by 0.1 at the end of the 15th epoch.
For DSVE-loc, we used the pretrained model posted on the original repository.
For the BERT model, we trained the embedding layers of both encoders for one epoch and the whole model for 29 epochs (i.e., 30 epochs in total).
The learning rate was initialized to $\alpha=2\times 10^{-4}$ and then multiplied by 0.1 at the ends of the first and 15th epochs.
The first 16 of the 24 layers of the BERT model were frozen in order to retain their pretrained features and to reduce memory consumption.
We used a data augmentation strategy similar to that of DSVE-loc; a random resize-and-crop to 256 pixels and a random horizontal flip were applied to the training images, and a resize to 350 pixels was applied to the validation and test images.

\subsection{Training of TOD-Net}
After the pretraining of each backbone model, we installed TOD-Net to the top of the backbone model.
For the conditional Real-NVP network, we set the number of coupling layers to 3, the number of hidden layers in a coupling layer to 2, and the number of hidden units to $2d=2048$ for VSE++ and BERT model and $d=2400$ for DSVE-loc.
We used ReLU activation functions~\cite{Nair2010}.
We trained TOD-Net for 30 epochs while freezing the feature extractors using Adam optimizer~\cite{Kingma2014b}.
The learning rate was initialized to $\alpha=2\times 10^{-5}$ and then multiplied by 0.1 at the end of the 15th epoch.

We report results averaged over three runs.
A typical measure of retrieval performance is recall at K, which is the fraction of the results that a positive target is ranked at the top $K$ candidates.
R@K denotes the performance of caption retrieval, and Ri@K denotes that of image retrieval.
Especially, R@1, R@5, R@10, Ri@1, Ri@5, Ri@10, and their average (mean rank; mR) have been commonly used for model evaluation.
After every epoch, the model was evaluated using mR for five folds of the validation set and the best snapshot was saved.
We omitted the median rank (Med r), which is the median rank of the first positive target, because it is saturated at 1.0 for all our methods and almost all existing state-of-the-art methods.

\section{Results and discussion}

\subsection{Combined with backbone models}
We combined TOD-Net with backbone models VSE++, DSVE-loc, and BERT model, and report the results in Table~\ref{tab:backbone}.
In all three cases, TOD-Net improves the retrieval performances R@1, R@5, Ri@1, and Ri@5 by significant margins.
In particular, despite the fact that DSVE-loc and BERT model are state-of-the-art methods, TOD-Net improves their performance.
Recall that the feature extractors are frozen when training TOD-Net.
This indicates that TOD-Net does not owe its performance improvements to longer training, but to the deformed embedding space that potentially dissolves the limitation of the fixed Euclidean space.
TOD-Net is considered to be universally applicable to visual-semantic embedding methods based on point embedding.
On the other hand, the improvements of R@10 and Ri@10 are limited.
When a backbone model extracts important concepts from an entity but fails in proper embedding, a positive target is ranked near but not at the top.
In this case, TOD-Net deforms the embedding space to adjust embedded vectors and ranks the positive target at the top, resulting in improvement of R@1 and Ri@1.
When a backbone model fails in concept extraction, a positive target is ranked far from the top, and TOD-Net cannot adjust it.
Hence, TOD-Net is good at improving R@1 and Ri@1 but not as effective at improving R@10 and Ri@10.

\begin{table}[t]\centering\small
  \caption{Results of TOD-Net Combined with Backbone Models}
  \label{tab:backbone}
  \tabcolsep=1mm
  \begin{tabular}{@{}llx{7mm}x{7mm}x{7mm}x{7mm}x{7mm}x{7mm}@{}}
    \toprule
    & \multicolumn{3}{c}{\textbf{Caption Retrieval}} & \multicolumn{3}{c}{\textbf{Image Retrieval}}\\
    \cmidrule(lr){2-4}\cmidrule(lr){5-7}
    \textbf{Similarity} & \textbf{R\!\!\:@\!\!\:1} & \textbf{R\!\!\:@\!\!\:5} & \textbf{R\!\!\:@\!\!\:10} &  \textbf{R\!\!\;i\!\!\:@\!\!\:1} & \textbf{R\!\!\;i\!\!\:@\!\!\:5} & \textbf{R\!\!\;i\!\!\:@\!\!\:10} & \textbf{mR} \\
    \midrule
    VSE++~\cite{Faghri2018} & 65.9 & 90.7 & 96.2 & 52.9 & 84.6 & 92.4 & 80.5 \\
    \ \ + TOD-Net   &\textbf{68.6} & \textbf{92.0} & \textbf{96.9} & \textbf{54.5} & \textbf{85.3} & 92.4   & \textbf{81.6} \\
    \midrule
    DSVE-loc~\cite{Engilberge2018}  & 69.8 & 91.9 & 96.6 & 55.9 & 86.9 & 94.0  & 82.5 \\
    \ \ + TOD-Net               & \textbf{72.3} & \textbf{93.4} & \textbf{97.4} & \textbf{58.5} & \textbf{88.3} & \textbf{94.6} & \textbf{84.1} \\
    \midrule
    BERT\!~\cite{Devlin2018} & 74.1 & 94.9 & 98.2 & 60.8 & 89.2 & 94.9 & 85.4 \\
    \ \ + TOD-Net            & \textbf{75.8} & \textbf{95.3} & \textbf{98.4} & \textbf{61.8} & \textbf{89.6} & \textbf{95.0} & \textbf{86.0}\\
    \bottomrule
  \end{tabular}\\
\end{table}

\begin{table}[t]
  \caption{Typical Successful Caption Retrieval by TOD-Net.}
  \label{tab:typical}
    \centering
    \tabcolsep=0mm
    \begin{tabular}{c@{\hspace*{5mm}}c}
      \includegraphics[width=30mm,bb=60 140 420 500,clip]{true_positive.jpg}&
      \includegraphics[width=30mm,bb=0 0 360 360,clip]{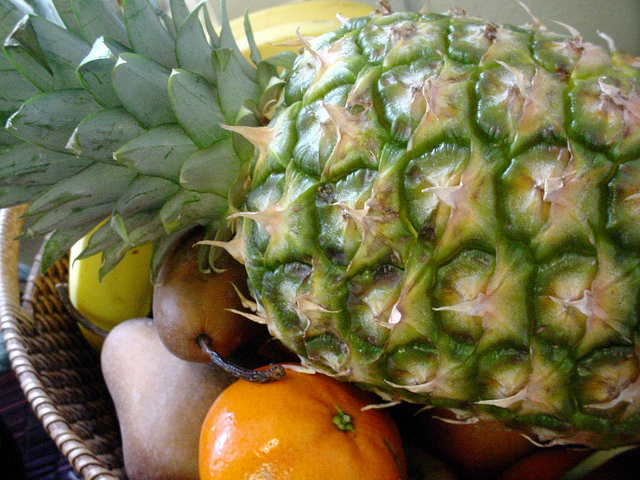}\\
      Query 1 & Query 2\\[2mm]
    \end{tabular}
    \footnotesize
    \begin{tabular}{lcl}
    \toprule
    \textbf{Model} & \multicolumn{2}{l}{\textbf{Top 5 retrieved captions} (\checkmark denotes a positive target)}\\
    \midrule
    &           & a person jumping a snow board in the air\\
    \multirow{2}{*}{VSE++}&\checkmark & \textcolor{red}{A man kite snowboarding on a sunny day}\\

    \multirow{2}{*}{(Query 1)}&           & A man on a snowboard does an air trick.\\
    &           & A snowboarder is is the air over the snow.\\
    &           & a person flying in the air while on a ski board.\\
    \midrule
    &\checkmark & \textcolor{red}{a person with green clothes and green board snowboarding.}\\
    VSE++
    &\checkmark & \textcolor{red}{A man kite snowboarding on a sunny day}\\
    + TOD-Net \hspace*{.1mm}
    &           & a person jumping a snow board in the air\\
    (Query 1)
    &\checkmark & \textcolor{red}{a person riding a snow board in the air}\\
    &           & A person flying through the air on the kite board.\\
    \cmidrule{1-3}\morecmidrules\cmidrule{1-3}
    &           & A lot of fruits that are in a bowl.\\
    \multirow{2}{*}{VSE++}
    &           & A fruit and vegetable stand has hanging fruit.\\
    \multirow{2}{*}{(Query 2)}
    &           & there are many crates filled with fruits and vegetable\\
    &           & A number of fruits and nuts on a stone\\
    &           & A pile of wooden boxes filled with fruits and vegetables.\\
    \midrule
    &\checkmark & \textcolor{red}{A pine apple on top of a pile of mixed fruit.}\\
    VSE++
    &           & A lot of fruits that are in a bowl.\\
    + TOD-Net \hspace*{.1mm}
    &\checkmark & \textcolor{red}{A bowl of assorted fruit with a huge pineapple on top.}\\
    (Query 2)
    &           & A number of fruits and nuts on a stone\\
    &\checkmark & \textcolor{red}{\scalebox{0.95}[1.0]{A fruit bowl containing a pineapple, an orange and several pears.}}\\
    \bottomrule
  \end{tabular}
\end{table}

\subsection{Interpretation of results}
In order to visualize the contribution of TOD-Net, we TOD-Net with VSE++.
We provide an example image (see Query 1 in Table~\ref{tab:typical}).
With the image as a query, the unmodified VSE++ did not rank a positive caption at the top in any of the three trials, while VSE++ with TOD-Net was successful in all three trials.
The unmodified VSE++ ranked captions such as ``a person jumping a snowboard in the air'' at the top.
These captions are apparently positive targets but not the optimal ones.
One of the actual positive captions is ``a person with green clothes and green board snowboarding;'' this caption focuses on the person's appearance.
Another positive caption is ``A man kite snowboarding on a sunny day;'' this caption describes the activity more specifically as kite snowboarding.
If the embedded vector of the image contains the concept relating to the appearance, it retrieves the former caption and fails to retrieve other captions that do not mention the appearance.
The embedded vector that focuses on the specific activity also retrieves nothing but the latter caption.
Hence, VSE++ extracts minimal concepts from the image to accept both cases, and thus retrieves many false positives.

Actually, VSE++ does not completely lose detailed concepts.
Through the deformation of the original embedded space $\mathcal Z$ as shown in Fig.~\ref{fig:deformation}, TOD-Net emphasizes the remaining specific concepts depending on the condition (i.e., the target).
As a result, a single image retrieves diverse captions as summarized in the lower portion of Table~\ref{tab:typical}.

Query 2 in Table~\ref{tab:typical} is another example.
The query image depicts a bowl full of fruit, among which the pineapple has the greatest presence.
To avoid a failure in retrieving a caption that does not mention the pineapple, the image encoder puts a little focus on the pineapple, which results in false positives that do not mention the pineapple.
Thanks to TOD-Net, which accepts a target caption as a condition, the image query can retrieve the positive captions that mention the pineapple.

A similar result can be found in the image retrieval, as shown in Table~\ref{tab:typical2}.
The query caption mentions the stairwell as well as the bench.
However, another caption of the same image instead mentions the artwork, and yet another one mentions the hallway.
Hence, it is inappropriate for their embedded vectors to focus on the stairwell and the hallway, as this leads to false positive images that depict only benches.
TOD-Net emphasizes the concept relating to the stairwell in both embedded vectors and retrieves the positive target selectively.

\begin{table}[t]
  \caption{Typical Successful Image Retrieval by TOD-Net.}
  \label{tab:typical2}
    \tabcolsep=0mm
    \footnotesize
    \begin{tabular}{lccccc}
      \multicolumn{6}{l}{\textbf{Query:}}\\
      \multicolumn{6}{l}{\hspace*{5mm}A wooden bench sits next to a stairwell.}\\
      \multicolumn{6}{l}{\textbf{Other captions of the same image:}}\\
      \multicolumn{6}{l}{\hspace*{5mm}A brown wooden bench sitting up against a wall.}\\
      \multicolumn{6}{l}{\hspace*{5mm}A bench next to a wall with a staircase behind it.}\\
      \multicolumn{6}{l}{\hspace*{5mm}Single wooden bench in corridor with artwork displayed above.}\\
      \multicolumn{6}{l}{\hspace*{5mm}The bench in the hallway of the building is empty.
      }\\
      \multicolumn{6}{l}{\textbf{Top 5 retrieved images} (A red border denotes the positive target): }\\
      \raisebox{4.5mm}{VSE++}&
      \includegraphics[width=1.25cm,height=1.25cm,cfbox=white 1pt 1pt]{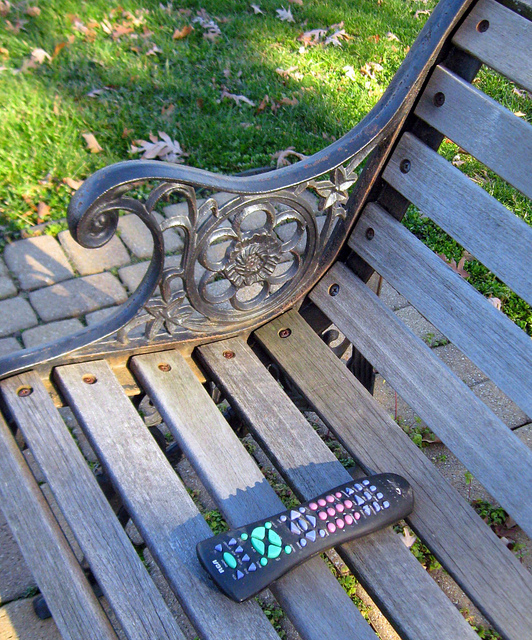}&
      \includegraphics[width=1.25cm,height=1.25cm,cfbox=white 1pt 1pt]{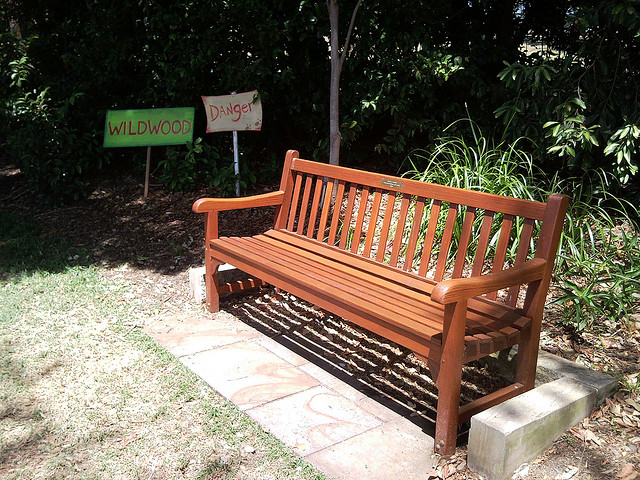}&
      \includegraphics[width=1.25cm,height=1.25cm,cfbox=white 1pt 1pt]{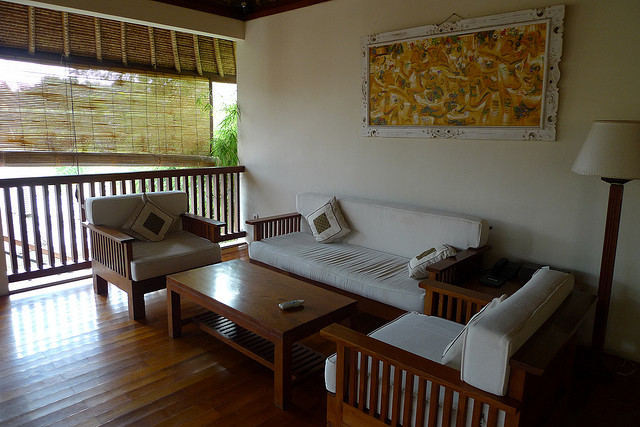}&
      \includegraphics[width=1.25cm,height=1.25cm,cfbox=white 1pt 1pt]{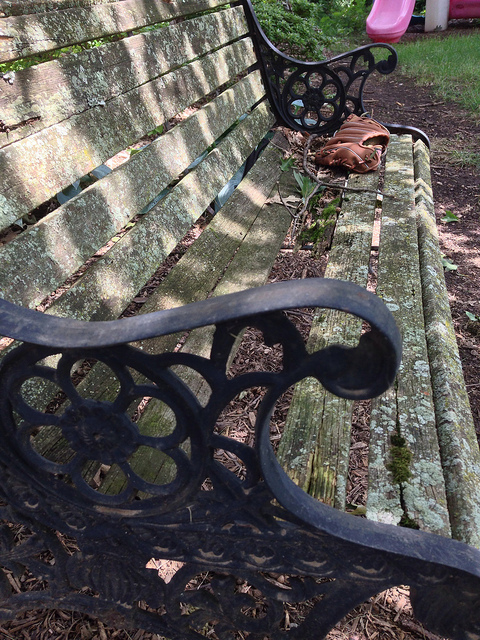}&
      \includegraphics[width=1.25cm,height=1.25cm,cfbox=white 1pt 1pt]{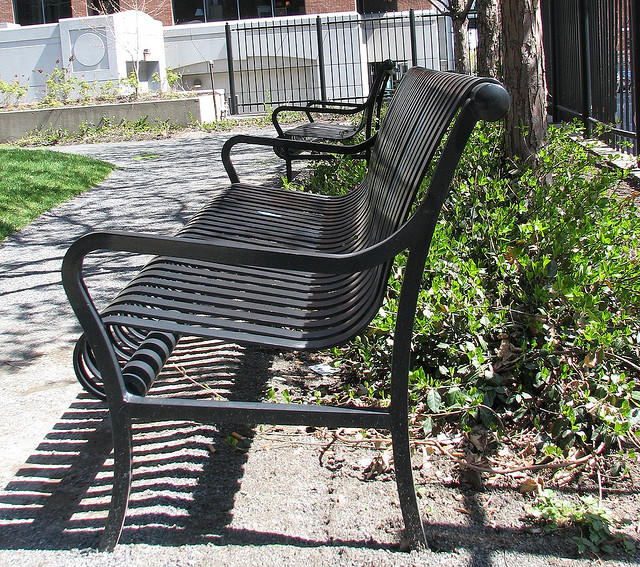}\\
      \raisebox{5mm}{
        \begin{tabular}{l}
          VSE++\\
          + TOD-Net\ \\
        \end{tabular}
      }&
      \includegraphics[width=1.25cm,height=1.25cm,cfbox=red 1pt 1pt]{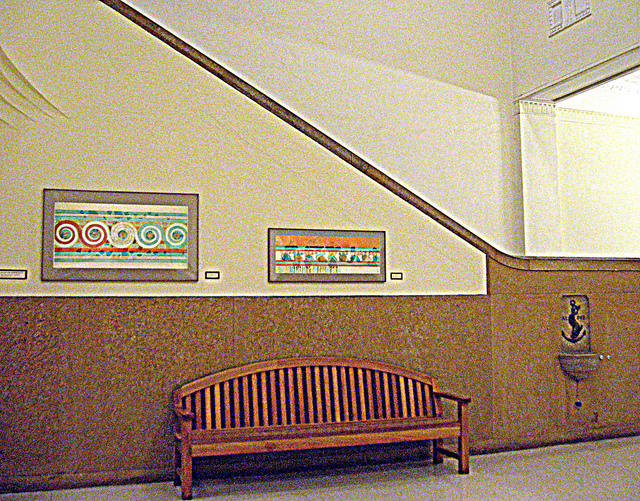}&
      \includegraphics[width=1.25cm,height=1.25cm,cfbox=white 1pt 1pt]{945.jpg}&
      \includegraphics[width=1.25cm,height=1.25cm,cfbox=white 1pt 1pt]{110.jpg}&
      \includegraphics[width=1.25cm,height=1.25cm,cfbox=white 1pt 1pt]{108.jpg}&
      \includegraphics[width=1.25cm,height=1.25cm,cfbox=white 1pt 1pt]{952.jpg}\\
    \end{tabular}
\end{table}

Recent studies on image-caption retrieval have employed cross-model attentions to pay attention to concepts shared by a query and a target~\cite{Huang2018f,Lee2018,Wang2019b,Liu2019,Lee2019,Huang2019b}.
Cross-modal attentions are performed at an early phase over multiple cropped regions of an image and words of a caption.
Conversely, TOD-Net is applied to the learned embedding space as the last step and retrieves diverse targets by a query.
The results indicate that the backbone models successfully extract detailed concepts from entities even when they are single-image encoders with neither object detectors nor cross-modal attentions.
The backbone models encounter a difficulty in the alignment of the entities in a single Euclidean space.
TOD-Net resolves the difficulty at a minimal modification.

\subsection{Ablation study}
We performed an ablation study on TOD-Net using VSE++.
We again report the performance of VSE++ with and without TOD-Net in the first two rows of Table~\ref{tab:ablation}.
In the third row, we report how TOD-Net performed without condition $c$.
Performance is slightly improved compared to the scenarios without TOD-Net in the second row, simply owing to a deeper network.
However, the improvement is limited, suggesting the importance of the condition.

As described in Section~\ref{sec:tod}, the condition $c$ of TOD-Net is a target.
A target is a caption $x$ for caption retrieval and an image $y$ for image retrieval.
Alternatively, one can consider other conditions such as a query.
One can use a caption $x$ as a condition $c$ for both caption and image retrieval, or use an image $y$.
We report these scenarios in the fourth through sixth rows of Table~\ref{tab:ablation}.
Performance is improved in many scenarios, but none of them is superior to the case in which condition $c$ is a target (see the first row).
This is TOD-Net determines which concept to emphasize easily with a target; meanwhile, it becomes more difficult to do it with other conditions.

\begin{table}[t]\centering\small
  \caption{Ablation Study of TOD-Net Combined with VSE++~\cite{Faghri2018}}
  \label{tab:ablation}
  \tabcolsep=0.5mm
  \hspace*{-1mm}
  \begin{tabular}{@{}llx{7mm}x{7mm}x{7mm}x{7mm}x{7mm}x{7mm}x{7mm}@{}}
    \toprule
    && \multicolumn{3}{c}{\textbf{Caption Retrieval}} & \multicolumn{3}{c}{\textbf{Image Retrieval}}\\
    \cmidrule(lr){3-5}\cmidrule(lr){6-8}
    \textbf{TOD-Net} & \hspace*{-.5mm}\textbf{Condition}\hspace*{-.5mm} & \textbf{R\!\!\:@\!\!\:1} & \textbf{R\!\!\:@\!\!\:5} & \textbf{R\!\!\:@\!\!\:10} &  \textbf{R\!\!\;i\!\!\:@\!\!\:1} & \textbf{R\!\!\;i\!\!\:@\!\!\:5} & \textbf{R\!\!\;i\!\!\:@\!\!\:10} & \textbf{mR} \\
    \midrule
    Real-NVP & target & \textbf{68.6} & \textbf{92.0} & \textbf{96.9} & \textbf{54.5} & \textbf{85.3} & 92.4 & \textbf{81.6} \\
    \midrule
    No       & ---               & 65.9 & 90.7 & 96.2 & 52.9 & 84.6 & 92.4 & 80.5 \\
    Real-NVP & no                & 66.6 & 91.5 & 96.8 & 54.0 & 85.1 & \textbf{92.6} & 81.1 \\
    \midrule
    Real-NVP & caption           & 68.5 & 91.5 & 96.7 & 53.6 & 84.7 & 91.9 & 81.2 \\
    Real-NVP & image             & 67.9 & 91.7 & 96.7 & 54.2 & 84.8 & 91.7 & 81.2 \\
    Real-NVP & query             & 68.2 & 91.8 & 96.8 & 53.8 & 84.8 & 92.1 & 81.3 \\
    \midrule
    MLP      & target            & 68.3 & 91.8 & 96.5 & 53.8 & 84.8 & 91.9 & 81.2 \\
    \bottomrule
  \end{tabular}\\
\end{table}

We also evaluate scenarios in which TOD-Net is composed of a conditional MLP (see the bottom row).
The MLP has four hidden layers, each having $2d$ units.
The performance is improved from the baseline but not superior to that of the conditional Real-NVP network.
We found the same results with two and six hidden layers.
The MLP approximates arbitrary functions while Real-NVP network approximates bijective functions.
Real-NVP network preserves the topological relations between embedded vectors in the original embedding space $\mathcal Z$, although the MLP could disturb it.

\subsection{Comparison with state-of-the-art models}
In Table~\ref{tab:sota}, we compare the experimental results against results from state-of-the-art methods that employ CNN-based single-image encoders.
TOD-Net with DSVE-loc achieved the best results on six of seven criteria, and TOD-Net with BERT outperformed all other methods on all seven criteria as emphasized by the bold text.

Several recent methods for image-caption retrieval employ an object detector and a cross-modal attention.
For example, SCO~\cite{Huang2018f}, SCAN~\cite{Lee2018}, and MTFN~\cite{Wang2019a} crop 10, 24, 36 regions, respectively, and merge them nonlinearly by cross-modal attentions.
It is known that a simple average over multiple regions significantly improves performance~\cite{Faghri2018}.
We summarize the results in Table~\ref{tab:objectdetection}\footnote{Note that MTFN~\cite{Wang2019a} also proposed a re-ranking algorithm, which finds the best match between a set of queries and a set of targets (not between a single query and a set of targets). We omit this result because the problem setting is completely different from those of the other studies. }; in each row, the number of cropped regions is indicated in parentheses.
Although TOD-Net takes only one region, its performance is already comparable or superior to those of the state-of-the-art methods based on object detectors.
In particular, TOD-Net with DSVE-loc outperforms methods in other studies in terms of Ri@1, and TOD-Net with the BERT model achieves the best results for all seven criteria.

Moreover, recent studies involving SCAN~\cite{Lee2018}, GVSE~\cite{Huang2019b}, and VSRN~\cite{Li2019a} reported the results of a two-model ensemble.
In this experimental setting, TOD-Net also improves performance significantly.
In particular, TOD-Net with the BERT model achieves the best results for all seven criteria.
TOD-Net with the DSVE-loc model achieves the second best results for five of seven criteria.

As in the previous section, these results also suggest that the cross-modal attention over multiple cropped regions is not the sole solution, but the adjustment of the embedding space works as a promising alternative.

\begin{table}[t]\centering\small
  \caption{Comparison with State-of-the-Art Methods using Single-Image Encoders}
  \label{tab:sota}
  \tabcolsep=1mm
  \begin{tabular}{@{}lx{7mm}x{7mm}x{7mm}x{7mm}x{7mm}x{7mm}x{7mm}@{}}
    \toprule
    & \multicolumn{3}{c}{\textbf{Caption Retrieval}} & \multicolumn{3}{c}{\textbf{Image Retrieval}}\\
    \cmidrule(lr){2-4}\cmidrule(lr){5-7}
    \textbf{Model}                           & \textbf{R\!\!\:@\!\!\:1} & \textbf{R\!\!\:@\!\!\:5} & \textbf{R\!\!\:@\!\!\:10} &  \textbf{R\!\!\;i\!\!\:@\!\!\:1} & \textbf{R\!\!\;i\!\!\:@\!\!\:5} & \textbf{R\!\!\;i\!\!\:@\!\!\:10} & \textbf{mR}\\
    \midrule
    m-CNN\cite{Ma2015b}              & 42.8 & 73.1 & 84.1 & 32.6 & 68.6 & 82.8 & 64.0 \\
    Order emb.\cite{Vendrov2015} & 46.7 & --   & 88.9 & 37.9 & --   & 85.9 & 64.9 \\
    DSPE+FV\cite{Wang2016c}          & 50.1 & 79.7 & 89.2 & 39.6 & 75.2 & 86.9 & 70.1 \\
    sm-LSTM\cite{Huang2016}          & 53.2 & 83.1 & 91.5 & 40.7 & 75.8 & 87.4 & 72.0 \\
    2WayNet\cite{Eisenschtat2017}    & 55.8 & 75.2 & --   & 39.7 & 63.3 & --   & -- \\
    DPC\cite{Zheng2017}              & 65.6 & 89.8 & 95.5 & 47.1 & 79.9 & 90.0& 78.0 \\
    VSE++\cite{Faghri2018}           & 64.6 & 90.0 & 95.7 & 52.0 & 84.3 & 92.0 & 79.8 \\
    GXN\cite{Gu2017}                 & 68.5 & --   & \underline{97.9} & 56.6 & --   & 94.5 & -- \\
    PVSE\cite{Song2019}          & 69.2 & 91.6 & 96.6 & 55.2 & 86.5 & 93.7 & 82.1 \\
    DSVE-loc\cite{Engilberge2018}    & 69.8 & 91.9 & 96.6 & 55.9 & 86.9 & 94.0 & 82.5 \\
    soDeep\cite{Engilberge2019}      & 71.5 & 92.8 & 97.1 & 56.2 & 87.0 & 94.3 & 83.2 \\
    \midrule
    \multicolumn{2}{l}{TOD-Net + (ours)}     & \\
    \ \ VSE++   &68.6 & 92.0 & 96.9 & 54.5 & 85.3 & 92.4   & 81.6 \\
    \ \ DSVE-loc               & \textbf{72.6} & \textbf{93.4} & 97.3 & \textbf{58.6} & \textbf{88.4} & \textbf{94.6} & \textbf{84.2} \\
    \ \ BERT {\footnotesize(1)}          & \textbf{75.8} & \textbf{95.3} & \textbf{98.4} & \textbf{61.8} & \textbf{89.6} & \textbf{95.0} & \textbf{86.0}\\
    \bottomrule
  \end{tabular}
\end{table}

\begin{table}[t]\centering\small
  \caption{Comparison with State-of-the-Art Methods using Object Detectors}
  \label{tab:objectdetection}
  \tabcolsep=1mm
  \begin{tabular}{@{}lx{6.4mm}x{6.4mm}x{6.4mm}x{6.4mm}x{6.4mm}x{6.4mm}x{6.4mm}@{}}
    \toprule
    & \multicolumn{3}{c}{\textbf{Caption Retrieval}} & \multicolumn{3}{c}{\textbf{Image Retrieval}}\\
    \cmidrule(lr){2-4}\cmidrule(lr){5-7}
    \textbf{Model}                & \textbf{R\!\!\:@\!\!\:1} & \textbf{R\!\!\:@\!\!\:5} & \textbf{R\!\!\:@\!\!\:10}  & \textbf{R\!\!\;i\!\!\:@\!\!\:1} & \hspace*{-.5mm}\textbf{R\!\!\;i\!\!\:@\!\!\:5} & \hspace*{-.5mm}\textbf{R\!\!\;i\!\!\:@\!\!\:10} & \textbf{m\!\!\:R} \\
    \midrule
    \ \ SCAN {\footnotesize t-i (24)~\cite{Lee2018}}      & 67.5 & 92.9 & 97.6 & 53.0 & 85.4 & 92.9 & 81.6 \\
    \ \ SCAN {\footnotesize i-t (24)~\cite{Lee2018}}      & 69.2 & 93.2 & 97.5 & 54.4 & 86.0 & 93.6 & 82.3 \\
    \ \ SCO {\footnotesize(10)~\cite{Huang2018f}}    & 69.9 & 92.9 & 97.5 & 56.7 & 87.5 & 94.8 & 83.2 \\
    \ \ R-SCAN {\footnotesize(36)~\cite{Lee2019}}    & 70.3 & 94.5 & 98.1 & 57.6 & 87.3 & 93.7 & 83.6 \\
    \ \ SGM {\footnotesize(36)~\cite{Wang2019d}}    &73.4 & 93.8 & 97.8 & 57.5 & 87.3 & 94.3 & 84.0\\
    \ \ MTFN {\footnotesize(36)~\cite{Wang2019a}}    & 71.9 & 94.2 & 97.9 & 57.3 & \underline{88.6} & \textbf{95.0} & 84.2 \\
    \ \ CAMP {\footnotesize(36)~\cite{Wang2019b}}    & 72.3 & 94.8 & \underline{98.3} & 58.5 & 87.9 & \textbf{95.0} & \underline{84.5} \\
    \ \ BFAN {\footnotesize($36$)~\cite{Liu2019}}    & \underline{73.7} & \underline{94.9} & -- & 58.3 &  87.5 & -- & --\\
    \midrule
    \multicolumn{2}{l}{TOD-Net + (ours)}     \\
    \ \ DSVE-loc {\footnotesize(1)}             & 72.6 & 93.4 & 97.3 & \textbf{58.6} & 88.4 & 94.6 & 84.2\\
    \ \ BERT {\footnotesize(1)}          & \textbf{75.8} & \textbf{95.3} & \textbf{98.4} & \textbf{61.8} & \textbf{89.6} & \textbf{95.0} & \textbf{86.0}\\
    \midrule
    \textbf{2-model ensemble} \\
    \midrule
    \ \ SCAN {\footnotesize($24\!\!\times\!\!2$)~\cite{Lee2018}}      & 72.7 & 94.8 & \underline{98.4} & 58.8 & 88.4 & 94.8 & 84.7 \\
    \ \ GVSE {\footnotesize($36\!\!+\!\!1$)~\cite{Huang2019b}} & 72.2 & 94.1 & 98.1 & 60.5 & 89.4 & \textbf{95.8} & 85.0 \\
    \ \ VSRN {\footnotesize($36\!\!\times\!\!2$)~\cite{Li2019a}}    & \underline{76.2} & 94.8 & 98.2 & \underline{62.8} & \underline{89.7} & 95.1 & \underline{86.1} \\
    \ \ BFAN {\footnotesize($36\!\!\times\!\!2$)~\cite{Liu2019}} & 74.9 & \underline{95.2} & -- & 59.4 & 88.4 & -- & -- \\
    \midrule
    \multicolumn{2}{l}{TOD-Net + (ours)}     \\
    \ \ DSVE-loc  {\footnotesize($1\!\!\times\!\!2$)}    & 75.4 & 94.4 & 97.8 & 60.9 & 89.6 & 95.3 & 85.6
    \\
    \ \ BERT {\footnotesize($1\!\!\times\!\!2$)}          & \textbf{78.1} & \textbf{96.0} & \textbf{98.6} & \textbf{63.6} & \textbf{90.6} & \textbf{95.8} & \textbf{87.1} \\
    \bottomrule
  \end{tabular}
\end{table}

\section{Conclusion}
In this paper, we have proposed TOD-Net, a novel module for embedding systems.
TOD-Net is installed on top of a pretrained embedding system and deforms the embedding space under a given condition.
Through the deformation, TOD-Net successfully emphasizes an entity-specific concept that is often denied owing to the diversity between entities belonging to the same group.
TOD-Net significantly outperforms state-of-the-art methods based on visual-semantic embedding for cross-modal retrieval on the MSCOCO dataset.
Moreover, despite the fact that TOD-Net takes only one region, it rivals or surpasses the performance of image-caption retrieval models based on object detectors.
Potential future research will focus on designing retrieval and translation systems that employ embedding internally.


\end{document}